\documentclass[letterpaper, 10 pt, conference]{ieeeconf}  

\IEEEoverridecommandlockouts                              

\usepackage{cite}
\usepackage{amsmath,amssymb,amsfonts}
\usepackage{algorithmic}
\usepackage{graphicx}
\usepackage{textcomp}
\usepackage{subfigure} 
\usepackage{setspace}
\usepackage{enumerate}
\usepackage{booktabs}
\usepackage{multirow}
\usepackage{float}
\usepackage{caption}
\usepackage{makecell}
\usepackage{xcolor}
\usepackage{array}
\usepackage{colortbl}
\definecolor{mygray}{gray}{.9}
\usepackage{ulem}

\usepackage{etoolbox}
\usepackage{balance}
\makeatletter
\patchcmd{\@makecaption}
  {\scshape}
  {}
  {}
  {}
\makeatletter
\patchcmd{\@makecaption}
  {\\}
  {.\ }
  {}
  {}
\makeatother

\usepackage{xcolor}

\title{\LARGE \bf
Learning Crowd Behaviors in Navigation with Attention-based Spatial-Temporal Graphs
}

\author{Yanying Zhou$^{1}$ and Jochen Garcke$^{1}$
\thanks{$^{1}$ All the authours are from Bonn University, Germany.}%
}

\begin{document}

\maketitle
\thispagestyle{empty}
\pagestyle{empty}

\begin{abstract}

Safe and efficient navigation in dynamic environments shared with humans remains an open and challenging task for mobile robots.
Previous works have shown the efficacy of using reinforcement learning frameworks to train policies for efficient navigation.
However, their performance deteriorates when crowd configurations change, i.e. become larger or more complex.
Thus, it is crucial to fully understand the complex, dynamic, and sophisticated interactions of the crowd resulting in proactive and foresighted behaviors for robot navigation.
In this paper, a novel deep graph learning architecture based on attention mechanisms is proposed, which leverages the spatial-temporal graph to enhance robot navigation. 
We employ spatial graphs to capture the current spatial interactions, and through the integration with RNN, the temporal graphs utilize past trajectory information to infer the future intentions of each agent.
The spatial-temporal graph reasoning ability allows the robot to better understand and interpret the relationships between agents over time and space, thereby making more informed decisions. 
Compared to previous state-of-the-art methods, our method demonstrates superior robustness in terms of safety,  efficiency, and generalization in various challenging scenarios.

\end{abstract}

\section{INTRODUCTION}
\label{sec:intro}

As artificial intelligence advances, robots are becoming essential in modern human life.
However, ensuring the safety and efficiency of social-aware robots navigating in crowds is a considerable challenge \cite{mavrogiannis2021core, moller2021survey, xiao2022motion}. 
These robots must not only avoid collisions and reach their destinations promptly but also act in a socially compliant manner and maintain a pleasant respectful interaction with pedestrians. 
Therefore, we aim to develop a proactive collision avoidance scheme that is both adaptive and generalizable.


Social-aware robot navigation tasks have been extensively studied and achieved many successful implementations.
Early rule-based methods \cite{van2008reciprocal, van2011reciprocal, 6698863,ferrer2017robot,truong2017toward} such as Optimal Reciprocal Collision Avoidance (ORCA) \cite{van2011reciprocal} and Social Forces (SFs) \cite{6698863,ferrer2017robot,truong2017toward} use artificially set geometric or physical rules, where the next action of the robot is taken depending on one-step rules and the current state.
This makes these methods short-sighted and highly dependent on the accuracy of the model.
Further, trajectory-based methods \cite{li2020socially, aoude2013probabilistically, chen2021interactive} plan an appropriate path for the robot after predicting the future trajectories of other agents, which results in far-sighted decisions.
However, both methods are prone to the freezing robot problem (FRP), which means that the planner determines that all feasible routes are unsafe and the robot gets stuck~\cite{5654369, sathyamoorthy2020frozone}.

\begin{figure}[t]
\setlength{\belowcaptionskip}{-6mm}
\centering
\includegraphics[width=\linewidth, height=0.65\linewidth]{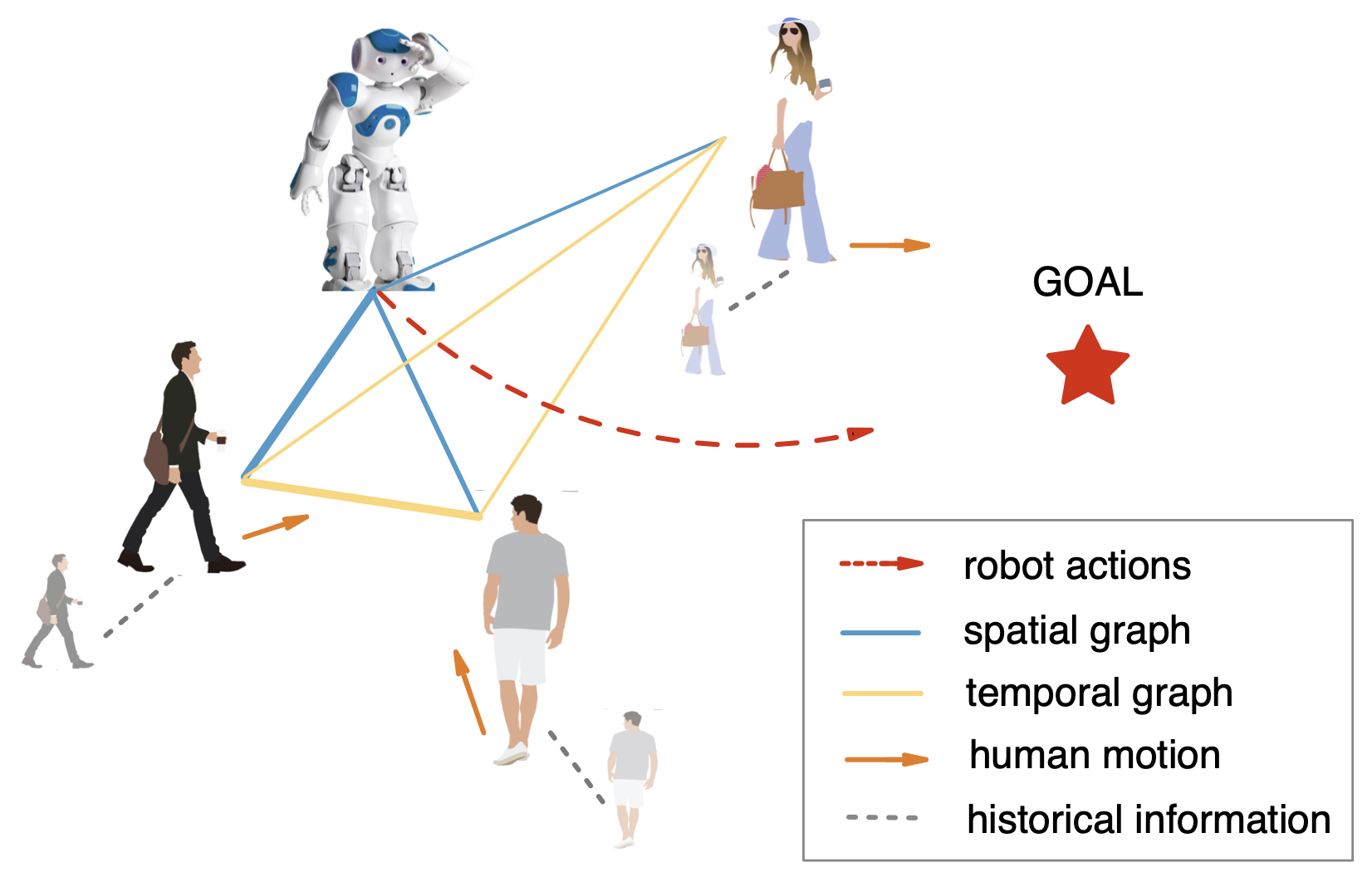}
\captionsetup{font={small}}
\caption{Illustration of our work. Our model uses spatial-temporal graphs to capture complex crowd dynamics, which aggregates both spatial and temporal attention maps from each agent.}
\label{fig:work}
\vspace{-2mm}
\end{figure} 

Recently, the use of Deep Reinforcement Learning (DRL) \cite{li2017deep, zhu2021deep, zhao2020sim, wang2022feedback, samsani2023memory, zeng2021robot} has enabled the development of learning-based methods for effective navigation strategies.
These methods train a value network using state transitions of agents to estimate the value of each state.
An optimal policy is then derived from the value network, guiding the robot to make decisions and choose the action that leads to the state with the highest value.
They implicitly encode the interactions between agents, capturing the collective impact of the crowd by incorporating pairwise interactions through LSTMs \cite{graves2012long} or the maximin operator.
However, interactions vary significantly across different environments, rendering navigation policies trained in one environment unsuitable for another. 
This is due to uncertainties in the dynamic crowd, which affects the way and impact of interactions. Specifically, as the density of crowds increases, it leads to increased mutual influence and interference among individuals, which in turn increases the difficulty and complexity of human-robot interaction in such environments.
Therefore, robots need to timely adjust their behavior strategies according to changes in crowd behavior, to better adapt to different scenarios and improve the quality of human-robot interaction.

In this paper, we propose an attention-based spatial-temporal graph learning framework for crowd navigation trained with DRL, which is called "ASTG".
First, we introduce graph attention networks to fully model both spatial and temporal relations of the crowd navigation scenario separately.
In addition, we incorporate an RNN \cite{medsker2001recurrent} to encode the historical trajectory information of the agents, before modeling temporal interactions for implicit reasoning about the intentions of all humans.
Then, we jointly aggregate the pairwise spatial-temporal interactions into a social attention mechanism to capture the relative importance of each human.
Together, a comprehensive understanding of crowd behaviors enables efficient robot navigation policies, adaptable to various crowded scenarios.

We present the following contributions:
(1) We introduce a novel deep graph neural network ASTG, which fully models the spatial and temporal relations efficiently in crowd navigation.
(2) We combine the above with a social-attention mechanism to encode the collective influence of neighbors. 
(3) We demonstrate the improved performance of our approach in comparison to previous methods, exhibiting strong robustness and generalization in both simple and complex scenarios with varying numbers of humans.


\section{RELATED WORK}

\subsection{Social-aware Robot Navigation}\label{subsec:social_nav}

Early works model agent interactions through \textit{rule-based methods} \cite{van2008reciprocal, van2011reciprocal, 6698863,ferrer2017robot,truong2017toward} or \textit{trajectory-based methods} \cite{li2020socially, aoude2013probabilistically, chen2021interactive}. Rule-based methods have a fast response, but are prone to encounter the freezing robot problem.
This is because they take actions that only consider the current state without a long-term view ahead.
Trajectory-based methods generate a path for the robot to follow after predicting other agents' future trajectories, which can be computationally intensive, especially when the environment has many dynamic agents.

\begin{figure*}[htbp]
\setlength{\abovecaptionskip}{-0.5mm}   
\setlength{\belowcaptionskip}{-3mm}
\centering
\includegraphics[width=0.9\linewidth]{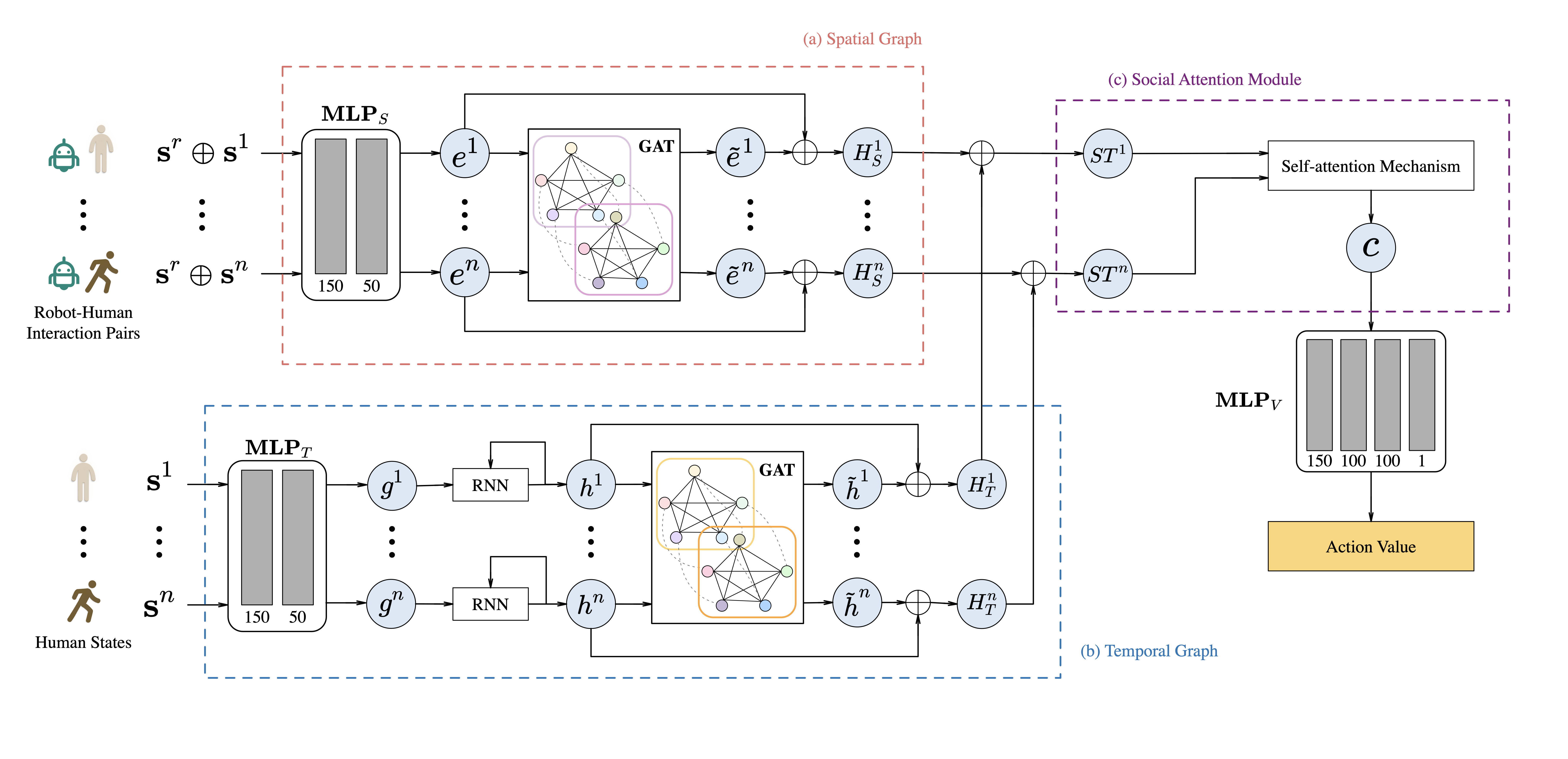}
\captionsetup{font={small}}
\caption{Network architecture from Section \ref{sec:method}. (a) The spatial graph utilizes the GAT to encode direct and indirect spatial interactions between agents. (b) The temporal graph incorporates an RNN to reason about the temporal interactions based on historical information. (c) The social attention module jointly aggregates the pairwise spatial-temporal interactions to capture the crowd representation in the crowd feature, which is then used to estimate the action values. }
\label{fig:astg}
\end{figure*}

Recently, combined with deep learning algorithms and RL, \textit{learning-based methods} \cite{chen2019crowd, liu2021decentralized,8202312, wang2023navistar, wang2022adaptive} learn how to achieve good navigation performance in an environment. 
A self-attention module is built to compute the relative importance in order to take actions, called SARL \cite{chen2019crowd}.
\cite{liu2021decentralized} propose a structural recurrent neural network to reason about spatial and temporal relationships for planning.

All of the above methods, however,  may not generalize well to new or unseen environments and may require additional training data and fine-tuning for each new environment. 
Significantly, deep reinforcement learning models can suffer from overfitting to the training data, leading to poor generalization performance in new states.
To tackle the problem, we introduce the graph attention network to adaptively weigh the importance of different nodes in the graph structure, providing a more informative and discriminative representation of the data.
In addition, the attention mechanisms in Graph Attention Network (GAT) can also provide additional regularization, helping to prevent overfitting to the training data and improve generalization performance.

\subsection{Graph-based Learning}\label{subsec:graph-learning}

Currently, graph neural networks (GNNs) are widely utilized in various applications, including social network analysis, to learn the relationships between entities \cite{he2023spatio, chen2020relational, chen2020robot, 9981037, zhou2022navigating, liu2022socially, zhang2020robot, zhou2022robot}. 
Combining the robust representation capabilities of graphs with the powerful end-to-end learning of neural networks, GNNs represent components as nodes and their relationships as edges. 
Through local message passing on graphs, these networks efficiently encapsulate relations between nodes, proving effective in a wide array of structured tasks.

GNNs can be used to extract efficient representations in crowd navigation.
Graph Convolutional Network (GCN) \cite{chen2020relational,chen2020robot,9981037} and GAT \cite{zhou2022robot, zhang2020robot, zhang2023robot} are the typical types of GNNs.
Chen et al. \cite{chen2020relational} present a relational graph model to learn the agents' interaction based on the two-layer GCN.
G-GCNRL \cite{chen2020robot} learns attention weights and aggregates the crowd information with two GCNs.
Unlike GCNs that use fixed weights in convolution operations, GAT uses learnable attention coefficients to weigh the contribution of each neighbor when aggregating information. 
In  \cite{zhou2022robot}, Zhou et al. introduce GAT to extract a nice graph representation.
However, it only refers to spatial interaction without the temporal dynamics, which also are useful for planning paths in crowd navigation.
Thus, our method applies GAT on the spatial graph and temporal graph separately to encode the environment relationships for dealing with uncertain dynamic environments.
In addition, for better decision-making, we incorporate RNN into encoding temporal features to reason about the future intentions of each agent based on the current and past observations of each human.
We show that combing spatial-temporal GAT with RNN improves performance in dynamic social navigation environments.


\section{PROBLEM FORMULATION}

\subsection{Crowd Navigation Modeling}\label{subsec:crowd_model}

We consider the crowd robot navigation task, where a holonomic robot needs to move towards a goal position through a crowd of $n$ humans. 
The task can be formulated as a sequential decision-making problem.
Assume that all agents move in a two-dimensional space with Euclidean geometry.
The current position $\mathbf{p} = [p_{x}, p_{y}]$, velocity $\mathbf{v} = [v_{x}, v_{y}]$ and radius $r$ of each agent can be observed by the rest.
The goal position $\mathbf{g} = [g_{x}, g_{y}]$ and the preferred velocity $v_{pref}$ are the hidden parts that the robot is aware of only for itself and one cannot observe the hidden states of humans.
Let $\mathbf{s}_{t}^{r}$ be the current state of the robot and $\mathbf{s}_{t}^{h} = [\mathbf{s}_{t}^{1}, \mathbf{s}_{t}^{2}, ..., \mathbf{s}_{t}^{n}]$ be the observable state of humans.
Thus, the joint state for robot navigation is $\mathbf{s}_{t}^{jn} = [\mathbf{s}_{t}^{r}, \mathbf{s}_{t}^{h}]$.

We transform the coordinate system with robot-centric coordinates, where the robot is set as the center and the $x$-axis is the direction towards the goal.
Thereby, the robot state and the $i$-th human state at time $t$ are defined as:
\begin{equation}
    \begin{split}
        & s^{r} = [d_g, v_{x}^{r}, v_{y}^{r}, v_{pref}, r],\\
        & s^{i} = [p_{x}^{i}, p_{y}^{i}, v_{x}^{i}, v_{y}^{i}, r^{i}, d^{i}, r^{i} + r],
    \end{split}
\end{equation}
where $d_{g}$ is the $L_2$ distance between the robot and its goal. $d^{i}$ is the $L_2$ distance between the robot and the $i$-th-human.

\subsection{Reinforcement Learning for Crowd Navigation}

In this work, a Deep V-learning approach~\cite{chen2020robot} is employed to evaluate the state values and choose the best action.
The robot starts from an initial state $\mathbf{s}^{jn}_{0}$ at the beginning of each episode.
Then, based on the learned policy $\pi (\mathbf{a}_{t}|\mathbf{s}^{jn}_{t})$, after taking an action $\mathbf{a}_{t} \in \mathcal{A}$ at each timestep $t$, the robot receives a reward $r_{t}$ and transitions to the successor state $\mathbf{s}^{jn}_{t+1}$.
For humans, based on their policies, they take actions and transition to their successor states.
For each episode, the process continues until the robot reaches the destination,  collides with other humans, or the navigation time $t$ exceeds the episode length $T$. The objective is to maximize the cumulative reward for each state and thereby to find the optimal policy $\pi^{*}(\mathbf{s}^{jn}_{t})$:
\begin{equation*}
\begin{split}
\pi^{*}&(\mathbf{s}^{jn}_{t}) = \mathop{\arg\!\max}_{\mathbf{a}_t} R (\mathbf{s}^{jn}_{t},\mathbf{a}_t) +  \\
& \gamma^{\Delta t \cdot v_{pref}}\int_{\mathbf{s}^{jn}_{t+\Delta t}}{P(\mathbf{s}^{jn}_{t},\mathbf{a}_t,\mathbf{s}^{jn}_{t+\Delta t})V^{\ast}(\mathbf{s}^{jn}_{t+\Delta t}) \, d\mathbf{s}^{jn}_{t+\Delta t}}, 
\end{split}
\end{equation*}
\begin{equation}
V^{*}(s_{t}^{jn}) = \sum_{t' = t}^{T} \gamma^{t' \cdot v_{pref}} R(\mathbf{s}^{jn}_{t'}, \pi^{*} (\mathbf{s}^{jn}_{t'})),
\end{equation}
where $\Delta t$ is the time step and $r_{t} = R (\mathbf{s}^{jn}_{t},\mathbf{a}_t)$ denotes the received reward at time $t$. $P(\mathbf{s}^{jn}_{t},\mathbf{a}_t,\mathbf{s}^{jn}_{t+\Delta t})$ is the transition probability between $t$ and $t+\Delta t$.
$\gamma \in (0,1)$ is a discount factor that is normalized by the preferred velocity $v_{pref}$. We utilize the reward function formulated as in \cite{chen2019crowd}, which provides rewards for task accomplishment while simultaneously imposing penalties for collisions or uncomfortable distances.


\section{METHODOLOGY}
\label{sec:method}

We propose a novel framework for social-aware robot navigation, see Fig.~\ref{fig:astg}.
This framework leverages the superior flexibility of graph structures over feed-forward neural networks, which are limited to a fixed number of agents.

\subsection{Spatial Graph Representation}

The spatial-temporal graph network calculates the spatial attention and temporal attention separately from the environment state matrix, which is used to represent the edges of the spatio-temporal graph.
In the spatial graph, we input rough pairwise spatial interactions to calculate agents' spatial dependencies.
For $i$-th human, we initially aggregate its state at the current moment with that of the robot to compute rough spatial interactions $(\mathbf{s}^{r} \oplus \mathbf{s}^{i})$. 
Then, we calculate spatial attention maps through graph attention network layers, representing the importance of adjacent interactions, and encode spatial features such as distance and relative direction between agents in pairwise interactions.
For instance, interaction at the $i$-th node is influenced by its connection with other nodes, with a more substantial influence reflected by a higher attention weight.
Simultaneously, this node aggregates spatial information from adjacent interactions through the influence of spatial edges, representing the collective impact on the node from the crowd dynamics.
This impact not only considers explicitly modeled pairwise human-robot interactions but also implicitly captures indirect human-human interactions and crowd-robot interactions through edge weights.

As shown in Fig. \ref{fig:astg}, in the spatial graph, we use the states of the robot and the $i$-th human at time $t$ to derive a node feature via a multilayer perceptron ($\text{MLP}_{S}$):
\begin{equation}
    e^{i}_{t} = f_{spatial}(\mathbf{s}^{r}_{t}, \mathbf{s}^{i}_{t};W_{e}),
\end{equation}
where $W_{e}$ are the network weights and $f_{spatial}(\cdot)$ is an MLP with a ReLU activation function. The rectified linear unit (ReLU) activation function used in the framework aims to capture non-linear features in the feedforward network.

We then use a graph attention layer to model spatial interactions in a crowd.
The input is denoted by $E = [e^{1}_{t},...,e^{n}_{t}]$ and $e^{i} = e^{i}_{t}$ for short.
The normalized spatial coefficients $\alpha_{ij}^{S}$ in the attention mechanism can be computed by:
\begin{equation}
    \alpha^{ij}_{S} = \frac{{\textrm{exp}}( {\textrm{LeakyRelu}}(a[\mathbf{W}e^{i}||\mathbf{W}e^{j}]))}{\sum_{k\in {\mathcal{N}}^{i}}{\textrm{exp}}( {\textrm{LeakyRelu}}(a[\mathbf{W}e^{i}||\mathbf{W}e^{k}]))},
\end{equation}
where $\alpha^{ij}_{S}$ represents the importance of node $j$ to node $i$.
$\mathbf{W}(\cdot)$ is the weight matrix and $||$ is the concatenation operation, while
${\mathcal{N}}^{i}$ indicates the neighborhood of node $i$ in the graph.
After passing through a fully connected network (FCN) $a(\cdot)$, a LeakyReLU activation follows.
Finally, a softmax function is used to normalize the spatial attention coefficients $\alpha_{ij}^{S}$. 
Thus, the output node features of the graph attention layer are:
\begin{equation}
    {\Tilde{e}}^{\,i} = \sigma (\sum_{j\in {\mathcal{N}}^{i}}\alpha^{ij}_{S}~\mathbf{W}e^{\,j}), \quad \Tilde{E} = [{\Tilde{e}}^{\,1},...,\Tilde{e}^{\,n}],
\end{equation}

In this work, through the graph attention layer, the $i$-th interaction feature is transformed into a higher-level spatial feature that models indirect robot-human and human-human interactions in a crowd. 
Additionally, residual connections are employed in the spatial and temporal graph networks to accelerate convergence and stabilize the framework \cite{he2016deep}.
Thus, the final spatial features $H_{spatial}$ is described by:
\begin{equation}
    H_{spatial} = E + {\Tilde{E}}, \quad H_{spatial} = [H_{S}^{1},...,H_{S}^{n}].
\end{equation}

\subsection{Temporal Graph Representation}

Due to the high motion-dependency of the temporal dimension, we introduce an RNN for each agent to capture the dynamics of its trajectory. 
Then, in the temporal graph, we treat the motion feature of each human as a node of the graph, and the edges signify the relationships between humans' motions, whose weights can indicate the importance level.
Since motion is continuous, we can predict future actions based on the current motion information. To achieve this, we utilize graph attention network layers to compute a temporal attention map, estimating the mutual influence of neighboring agents' motion behaviors and aggregating these influences to form temporal features. 
These temporal features encompass both the current trajectory information of humans and predictions of future trajectories, allowing the robot to better understand human behavior and intentions.

First, we embed $i$-th human's state through an $\text{MLP}_{T}$ into a fixed length vector $g^{i}_{t}$:
\begin{equation}
    g^{i}_{t} = f_{temporal}(\mathbf{s}^{i}_{t};W_{g}),
\end{equation}
where $W_{g}$ are the embedding weights and $f_{temporal}(\cdot)$ is an MLP with ReLU activation function. Then, we process $g_{i}^{t}$ with an RNN cell:
\begin{equation}
    h^{i}_{t} = {\textrm{RNN}}(h^{i}_{t-1},g^{i}_{t}),
\end{equation}
where $h^{i}_{t}$ is the hidden state at time $t$, which changes over time, reflecting the evolution of $i$-th agent's state.

The hidden states $h^{i}_{t}$ are then fed into the graph attention layer to model temporal interactions.
GAT learns the attention distribution between nodes adaptively, enabling it to weigh the information from different nodes based on their relationship. 
This implies that GAT facilitates the propagation of temporal interaction information between nodes.
The input is $H = [h^{1}_{t},...,h^{n}_{t}]$ and $h^{i} = h^{i}_{t}$ for short.
The normalized temporal coefficients $\alpha_{ij}^{T}$ is described by:
\begin{equation}
    \alpha^{ij}_{T} = \frac{{\textrm{exp}}( {\textrm{LeakyRelu}}(\alpha[\mathbf{W}h^{i}||\mathbf{W}h^{j}]))}{\sum_{k\in {\mathcal{N}}^{i}}{\textrm{exp}}( {\textrm{LeakyRelu}}(\alpha[\mathbf{W}h^{i}||\mathbf{W}h^{k}]))},
\end{equation} 

Combining RNN and GAT allows for a more accurate capture of the evolving states and interactive changes of agents in a social environment over time. 
While RNN captures the temporal variations in agents' states, GAT, through its attention mechanism, considers interactions with other nodes, leading to a finer-grained temporal dynamic modeling.
By comprehensively considering the evolution of each agent's state alongside their social relationships, the robot can better predict future human behaviors and intentions, and even assist in analyzing collective behavioral patterns within a group.
Hence, the final temporal graph feature with a one-layer GAT can be described as follows:
\begin{equation}
    \Tilde{h}^{i} = \sigma (\sum_{j\in {\mathcal{N}}^{i}}\alpha^{ij}_{T}~\mathbf{W}h^{j}), \quad \Tilde{H} = [\Tilde{h}^{1},...,\Tilde{h}^{n}],
\end{equation} 
After deploying a residual connection structure, the final temporal features $H^{temporal}$ is:
\begin{equation}
    H_{temporal} = H + \Tilde{H}, \quad H_{temporal} = [H_{T}^{1},...,H_{T}^{n}].
\end{equation} 

\subsection{Social Attention Mechanism} 

To capture the uncertainty of crowd movements, we build a social attention module that fuses the spatial and temporal features of each agent and captures dependencies among agents. 
Inspired by \cite{chen2019crowd}, we build a self-attention network to assign attention weights $w^{i}$ for each agent's aggregated spatial-temporal pairwise feature $ST^{\,i} = [H_{S}^{i}, H_{T}^{i}]$ and encode the collective impact of a crowd. 

First, $ST^{\,i}$ is transformed into an attention score $w^{i}$: 
\begin{equation}
    ST^{\,m} = \frac{1}{n}\sum_{k=1}^{n}ST^{\,k}, \quad
    w^{i} = f_{\alpha}(ST^{\,i}, ST^{\,m}; W_{\alpha}),
\end{equation} 
where $W_{\alpha}$ is the weight and $f_{\alpha}$ is nonlinear transformation.
We then obtain the final crowd representation $c$ by the product sum of each spatial-temporal pairwise feature $ST^{\,i}$ and attention weights $w_{i}$:
\begin{equation}
    c = \sum_{i=1}^{n} softmax(w^{i})ST^{\,i}
\end{equation} 

\subsection{Graph-Based Planning}

The crowd representation $c$ is the input to $\text{MLP}_{V}$ to estimate the state value $V$ for planning:
\begin{equation}
    V = f_{v}(\mathbf{s}^{r}, c; W_{v})
\end{equation}
where $W_{v}$ denotes the weights.

\begin{figure*}[htbp]
\setlength{\abovecaptionskip}{-0.5mm}   
\setlength{\belowcaptionskip}{-4mm}
\centering
\subfigure
{
\begin{minipage}[b]{0.16\linewidth}
\centering
\includegraphics[width=\linewidth]{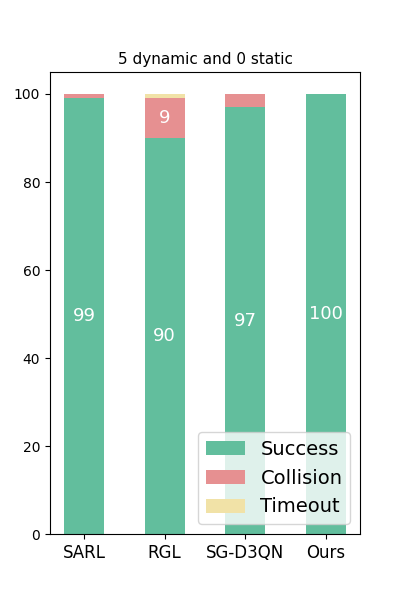}
\end{minipage}%
\begin{minipage}[b]{0.16\linewidth}
\centering
\includegraphics[width=\linewidth]{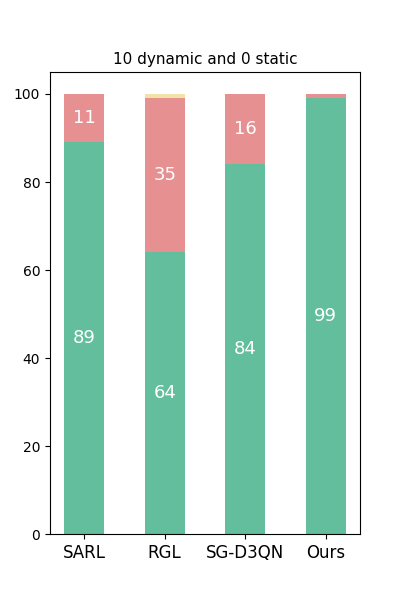}
\end{minipage}%
\begin{minipage}[b]{0.16\linewidth}
\centering
\includegraphics[width=\linewidth]{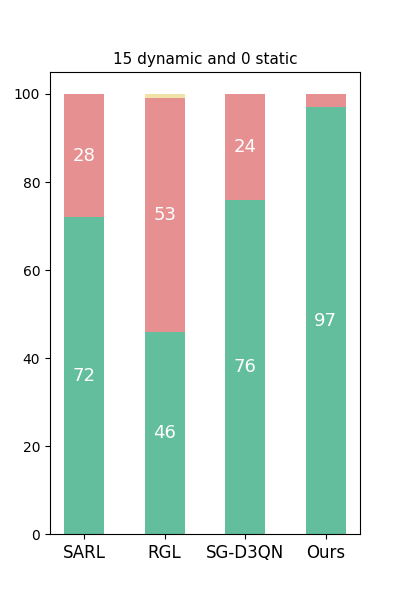}
\end{minipage}%
\begin{minipage}[b]{0.16\linewidth}
\centering
\includegraphics[width=\linewidth]{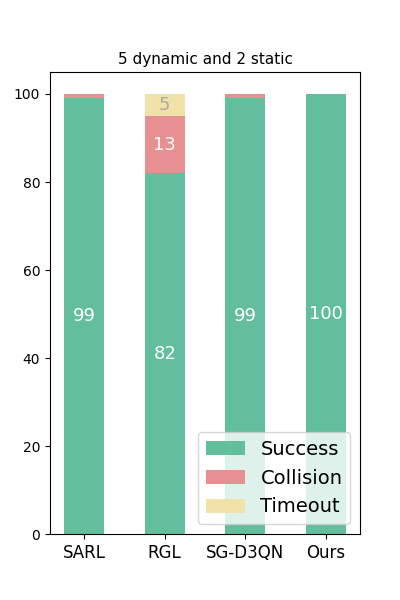}
\end{minipage}%
\begin{minipage}[b]{0.16\linewidth}
\centering
\includegraphics[width=\linewidth]{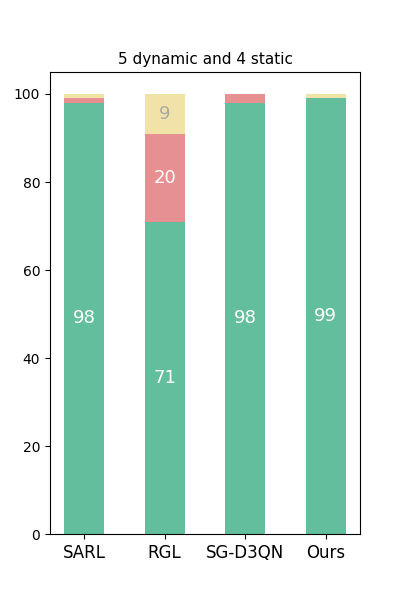}
\end{minipage}%
\begin{minipage}[b]{0.16\linewidth}
\centering
\includegraphics[width=\linewidth]{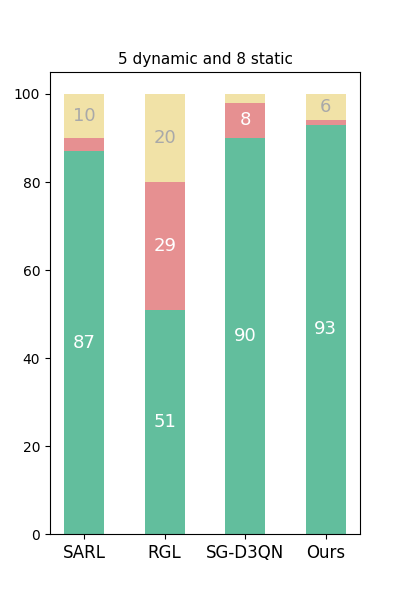}
\end{minipage}%
}%
\captionsetup{font={small}}
\caption{Quantitative evaluation on scenarios with different numbers of humans.}
\label{fig:simple}
\vspace{2mm}
\end{figure*}

\begin{table*}[htbp]
\setlength{\belowcaptionskip}{-6mm}
    \renewcommand\arraystretch{1.2}
	\centering
	\captionsetup{font={normalsize}}
    \resizebox{\linewidth}{!}{
    \begin{tabular}{l|ccc|ccc|ccc|ccc|ccc|ccc}
    \hline
    \multirow{3}*{\makecell{    Methods}}&
    \multicolumn{9}{c|}{different numbers of dynamic humans}&\multicolumn{9}{c}{5 dynamic humans and different numbers of static humans}\\
    \cline{2-19}
& \multicolumn{3}{c|}{Disc.(\%)}& \multicolumn{3}{c|}{$t_{succ.nav}$}& \multicolumn{3}{c|}{$t_{weighted.nav}$}& \multicolumn{3}{c|}{Disc.(\%)}& \multicolumn{3}{c|}{$t_{succ.nav}$}& \multicolumn{3}{c}{$t_{weighted.nav}$}\\
    \cline{2-19}
    & 5& 10&15& 5& 10&15& 5& 10&15&2&4&8&2&4&8&2&4&8\\
    \hline    SARL~\cite{chen2019crowd}&0.02&0.11&0.22&10.58&11.64&\textbf{12.33}&10.91&14.18&17.80&0.05&0.09&0.15&10.51&\textbf{10.64}&\textbf{11.16}&11.10&11.71&13.52\\
    RGL~\cite{chen2020relational}&0.21&0.42&0.53&10.23&11.43&12.51&12.57&17.65&21.60&0.34&0.43&0.57&10.59&10.97&12.06&13.88&15.53&18.41\\
    SG-DQN~\cite{zhou2022robot}&0.05&0.08&0.09&\textbf{10.03}&\textbf{11.24}&13.41&\textbf{10.63}&13.71&16.56&0.04&0.06&\textbf{0.09}&\textbf{10.07}&11.24&12.07&\textbf{10.42}&11.79&13.62\\
    Ours&\textbf{0.01}&\textbf{0.04}&\textbf{0.08}&11.40&12.65&13.28&11.45&\textbf{12.97}&\textbf{14.13}&\textbf{0.03}&\textbf{0.06}&0.10&11.09&11.22&11.97&11.25&\textbf{11.56}&\textbf{12.83}\\
    \hline
    \end{tabular}
    }
	\captionsetup{font={small}}
  	\caption{Evaluation performance comparison in the simple scenarios. "Disc. (\%)" is the average frequency of duration that the robot invades the humans' comfort area. "$t_{succ.nav}$" is the average navigation time of success cases. "$t_{weighted.nav}$" (Eq.(18)) is a novel weighted navigation time metric considering the impact of the discomfort steps and collision cases.}
	\label{tab1}
\end{table*}

\section{EXPERIMENTS}

\subsection{Simulation Environment}

We build our 2D simulation environment for crowd robot navigation following \cite{chen2019crowd}.
We keep the invisible setting for the robot to prevent humans from overreacting to the robot and thus avoid learning an aggressive navigation policy that moves too directly to the goal position and gets too close to humans.
We use circle crossing scenarios with a radius of 4m in our simulation.
All humans are controlled by the ORCA policy.
Dynamic humans are randomly positioned on the circle and will move to opposite positions on the same circle, while static humans are randomly located in the same circle and do not move.
To fully analyse the effectiveness of the proposed model, we evaluate all models for 1000 random test cases in both simple and complex scenarios. 

\begin{figure}[t]
\setlength{\abovecaptionskip}{-0.5mm}   
\setlength{\belowcaptionskip}{-6mm}
\centering
\subfigure[DS (distributed)]
{
\begin{minipage}[b]{0.30\linewidth}
\centering
\includegraphics[width=\linewidth]{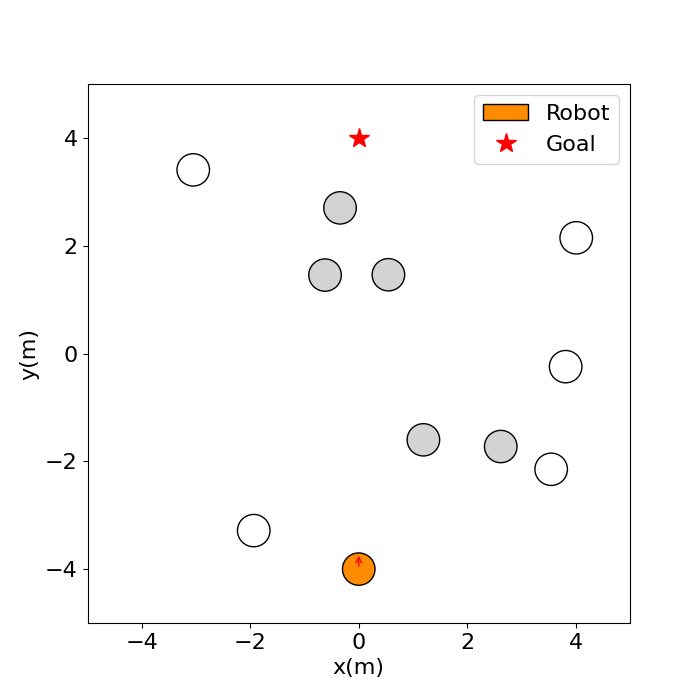}
\end{minipage}%
}
\subfigure[RO (row3and2)]
{
\begin{minipage}[b]{0.30\linewidth}
\centering
\includegraphics[width=\linewidth]{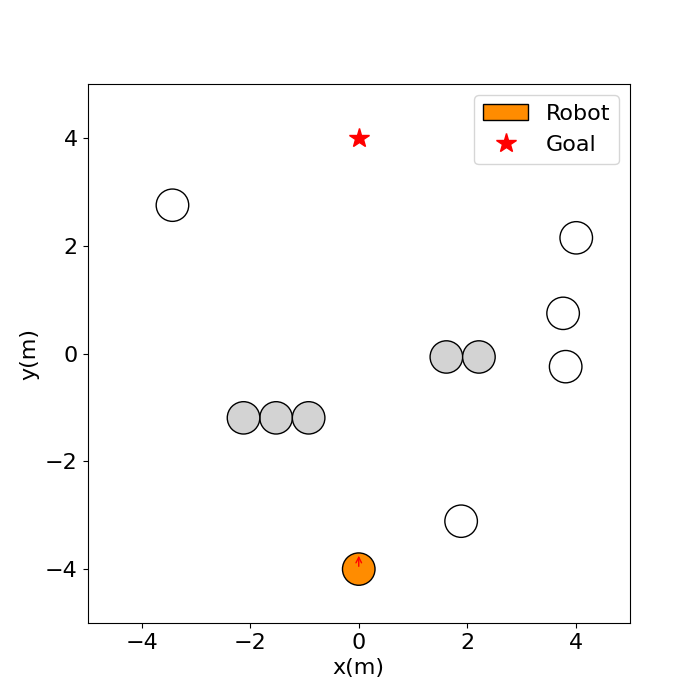}
\end{minipage}%
}
\subfigure[CO (concave5)]
{
\begin{minipage}[b]{0.30\linewidth}
\centering
\includegraphics[width=\linewidth]{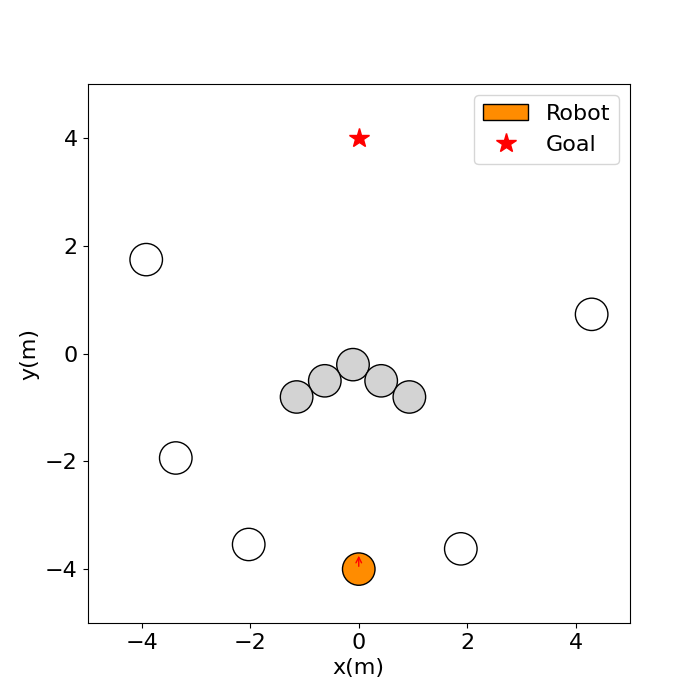}
\end{minipage}%
}%
\captionsetup{font={small}}
\caption{Simple and complex scenarios. They are all with 5 dynamic humans and 5 static humans with different combinations, such as (a) Completely dispersed. (b) Divided into two rows with 3 and 2 static humans separately. (c) Formed as concave.}
\label{fig:env}
\end{figure}

\subsection{Implementation Details}

\textbf{\textit{Baselines for comparison:}}
We compare the performance of our model with three state-of-the-art methods: SARL \cite{chen2019crowd}, RGL \cite{chen2020relational} and SG-DQN \cite{zhou2022robot}. 
SARL is a baseline for Deep V-learning.
In addition, RGL and SG-DQN are used as baselines for graph-based learning methods.

\textbf{\textit{Training:}} We train all methods in an environment that consists of 5 dynamic and 2 scattered static humans using data from 10k episodes. 
For a fair comparison, the network architectures of all baselines are trained as stated in the original papers.

\textbf{\textit{Evaluation:}}
We evaluate the generalization capability of the policy of each method, learned in the above training scenario, in both simple and complex scenarios with 1k episodes. To vary the start and end positions of agents over the episodes we use different random seeds, where each (random) episode configuration is evaluated for all methods. 
As evaluation metrics, we use the percentage of success, collision, and timeout cases, respectively, as well as  navigation time-related metrics.

Simple scenarios are set up with different numbers of distributed dynamic and/or static humans (like Fig. \ref{fig:env}(a)), such as scenarios with only $n = 5, 10, 15$ dynamic humans, and scenarios with 5 dynamic humans or $m = 2, 4, 8$ scattered static humans.
In addition, based on 5 dynamic humans, the complex scenario has 5 static humans with different group combinations (like Fig. \ref{fig:env}(b),(c)), whose positions are randomly set in the circle.

 	

\subsection{Quantitative Evaluation}

\begin{figure}[t]
\setlength{\abovecaptionskip}{-0.5mm}   
\setlength{\belowcaptionskip}{-3mm}
\centering
\subfigure
{
\begin{minipage}[b]{0.25\linewidth}
\centering
\includegraphics[width=\linewidth]{images/dynamic10static0.png}
\end{minipage}%
\begin{minipage}[b]{0.25\linewidth}
\centering
\includegraphics[width=\linewidth]{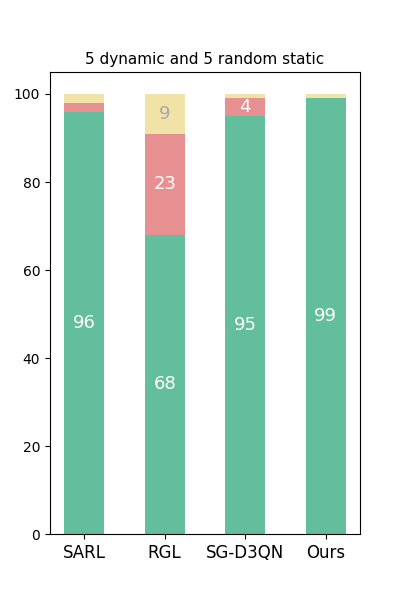}
\end{minipage}%
\begin{minipage}[b]{0.25\linewidth}
\centering
\includegraphics[width=\linewidth]{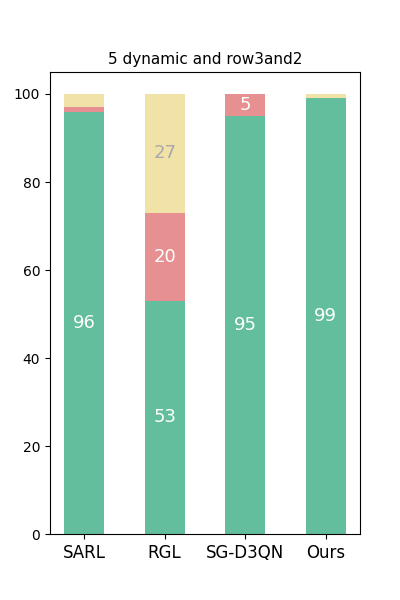}
\end{minipage}%
\begin{minipage}[b]{0.25\linewidth}
\centering
\includegraphics[width=\linewidth]{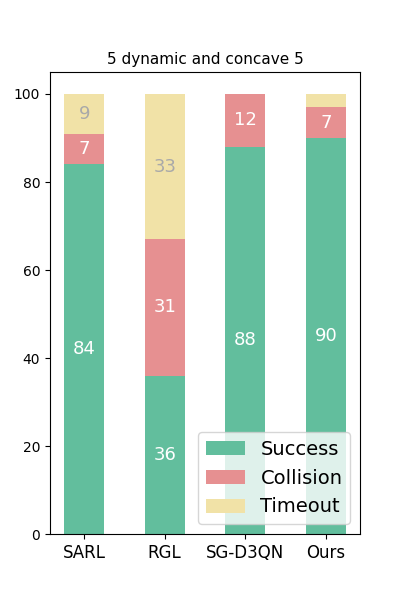}
\end{minipage}%
}%
\captionsetup{font={small}}
\caption{Quantitative evaluation under complex scenarios.}
\label{fig:complex}
\end{figure}

To show the effectiveness of our method, we compare our method with SARL \cite{chen2019crowd}, RGL \cite{chen2020relational} and SG-DQN \cite{zhou2022robot}. Fig.\ref{fig:simple}, \ref{fig:complex}, Tab.\ref{tab1}, \ref{tab3} show the results of all methods under different environments, where ASTG is the best one with the highest success rate and the lowest comfort rates in all scenarios. 

In Fig. \ref{fig:simple} and \ref{fig:complex}, we observe that the SARL method \cite{chen2019crowd} demonstrates restricted performance and social compliance due to its inability to fully capture intricate social interactions.
Moreover, despite utilizing GNN for modeling human-robot interactions, RGL \cite{chen2020relational} has not demonstrated strong generalization performance. 
This limitation stems from the GCN’s simple convolution operations applied on features dictated by fixed rule-based weights, which weaken adaptability to varying environments, leading to a sharp increase in collision rates as the obstacles' amount grows. 
In contrast, GAT can assign different weights based on the importance of these features, thereby better capturing critical environmental characteristics.
This advantage is visible in SG-DQN’s \cite{zhou2022robot} competitive stance alongside our ASTG. 
However, SG-DQN \cite{zhou2022robot} lacks the ability to capture time-varying motion features, thus reducing predictive accuracy. 
In comparison, our ASTG exhibits significantly slower increases in collision rates and navigation times in pure dynamic environments and environments containing both dynamic and static humans, demonstrating excellent generalization performance.

\begin{table}[t]
\setlength{\belowcaptionskip}{-6mm}
    \renewcommand\arraystretch{1.2}
	\centering
	\captionsetup{font={normalsize}}
 	
    \resizebox{\linewidth}{!}{
    \begin{tabular}{l|ccc|ccc|ccc}
    \hline
    \multirow{2}*{\makecell{Group \\ statics}}& \multicolumn{3}{c|}{Disc.(\%)}& \multicolumn{3}{c|}{$t_{succ.nav}$}&  \multicolumn{3}{c}{$t_{weighted.nav}$}\\
    \cline{2-10}
    &DS&RO&CO&DS&RO&CO&DS&RO&CO\\
    \hline
    SARL~\cite{chen2019crowd}&0.11&0.10&0.05&\textbf{10.83}&\textbf{11.08}&11.54&12.18&12.31&13.21\\
    RGL~\cite{chen2020relational}&0.44&0.56&0.59&11.25&11.37&11.27&16.22&16.50&18.58\\
    SG-DQN~\cite{zhou2022robot}&0.07&0.07&0.06&10.96&11.25&11.15&11.93&12.20&13.06\\
    Ours&\textbf{0.06}&\textbf{0.06}&\textbf{0.03}&11.35&11.34&\textbf{11.05}&\textbf{11.75}&\textbf{11.76}&\textbf{12.21}\\
    \hline
    \end{tabular}
    }
    \captionsetup{font={small}}
    \caption{Evaluation performance comparison in the scenarios with 5 dynamic humans and different static groups (DS (distributed), RO (row3and2), and CO (concave)). }
	\label{tab3}
\end{table}

From Tab.\ref{tab1}, \ref{tab3}, it shows that the baselines enjoy a better performance in $t_{succ.nav}$, which is defined as the average navigation time of the successful cases in seconds.
However, they have either a higher discomfort frequency or higher collision rate, which means, to achieve a shorter navigation time, the robot takes a lot of aggressive actions, badly affecting human comfort or even having collisions.
To better measure and analyse the behavior, we propose a weighted navigation time metric 
that penalizes aggressive policies:
\begin{equation}
    t_{weighted.nav} = \frac{t_{dnav}+N_{coll}*t_{limit}}{N_{succ}+N_{coll}},
\end{equation}
where $N_{succ}$ and $N_{coll}$ are the number of success and collision cases, respectively. 
$t_{limit}$ is the maximum navigation time, which set as 25s in our simulations. 
$t_{dnav}$ is extended from the original navigation time $t_{nav}$ to penalize the impact of discomfort. 
Specifically, if a step is causing discomfort, we add  half of a time step to $t_{succ.nav}$, which equals $t_{dnav} = t_{succ.nav}+N_{disc}*0.5*{\Delta}t$ and $N_{disc}$ is the number of discomfort steps in success cases.
Thus, when considering collision cases and the discomfort frequency, our method shows a shorter weighted navigation time $t_{weighted.nav}$. 


\begin{figure}[t]
\setlength{\abovecaptionskip}{-0.5mm}   
\setlength{\belowcaptionskip}{-3mm}
\centering
\subfigure[Only spatial graph]
{
\begin{minipage}[b]{0.31\linewidth}
\centering
\includegraphics[width=\linewidth]{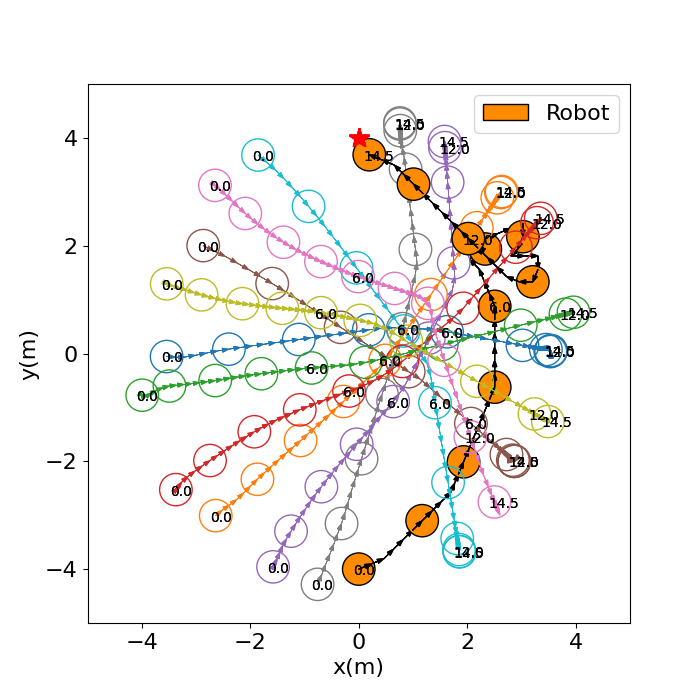}
\end{minipage}%
}
\subfigure[Only temporal graph]{
\begin{minipage}[b]{0.31\linewidth}
\centering
\includegraphics[width=\linewidth]{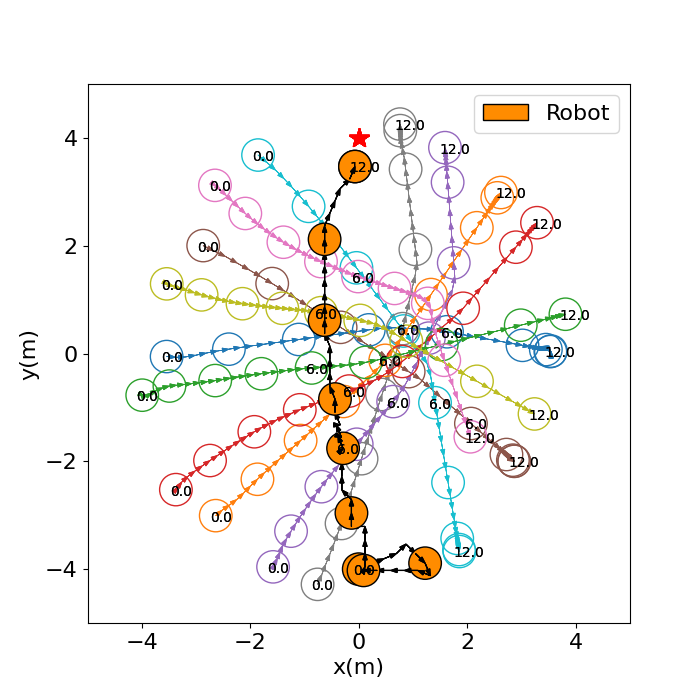}
\end{minipage}%
dyn}
\subfigure[Full ASTG]{
\begin{minipage}[b]{0.31\linewidth}
\centering
\includegraphics[width=\linewidth]{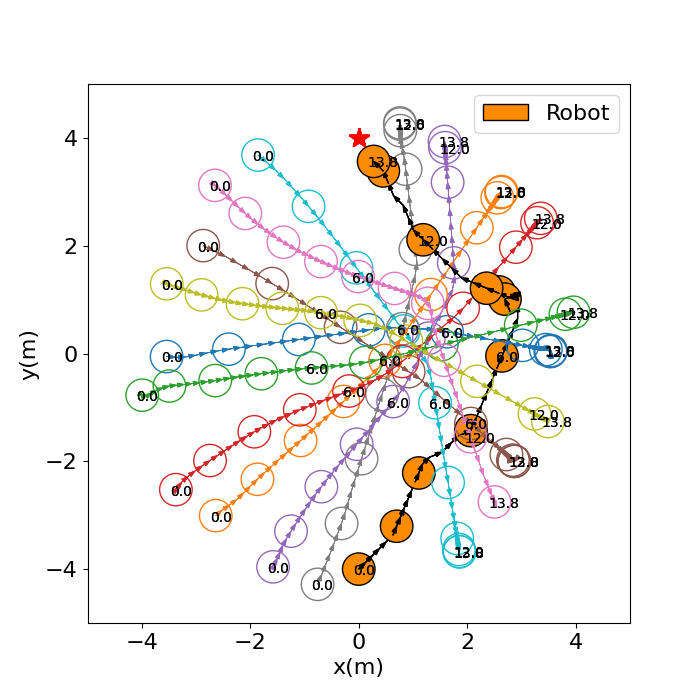}
\end{minipage}%
}%
\captionsetup{font={small}}
\caption{Simulation trajectories on the same case.}
\label{fig:ablation}
\end{figure}

\begin{figure}[t]
\setlength{\abovecaptionskip}{-0.5mm}   
\setlength{\belowcaptionskip}{-3mm}
\centering
\subfigure[SARL]
{
\begin{minipage}[b]{0.31\linewidth}
\centering
\includegraphics[width=\linewidth]{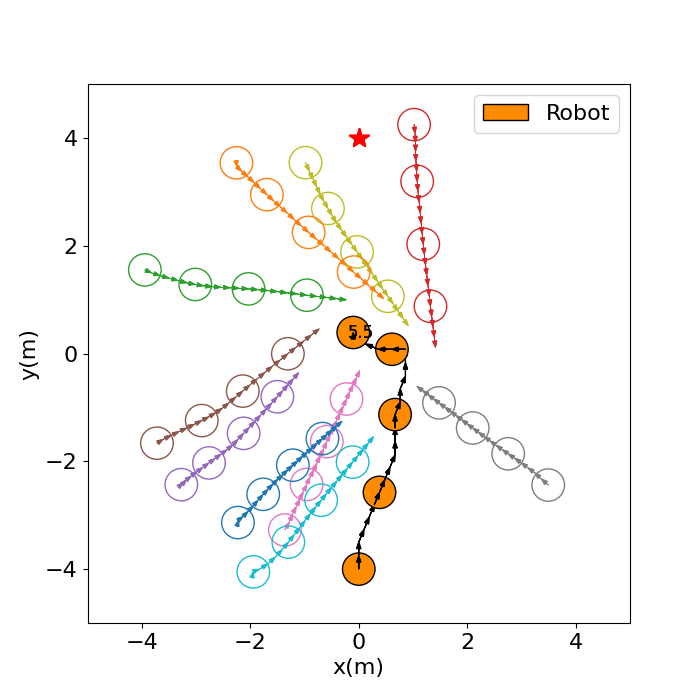}
\end{minipage}%
}
\subfigure[SG-DQN]{
\begin{minipage}[b]{0.31\linewidth}
\centering
\includegraphics[width=\linewidth]{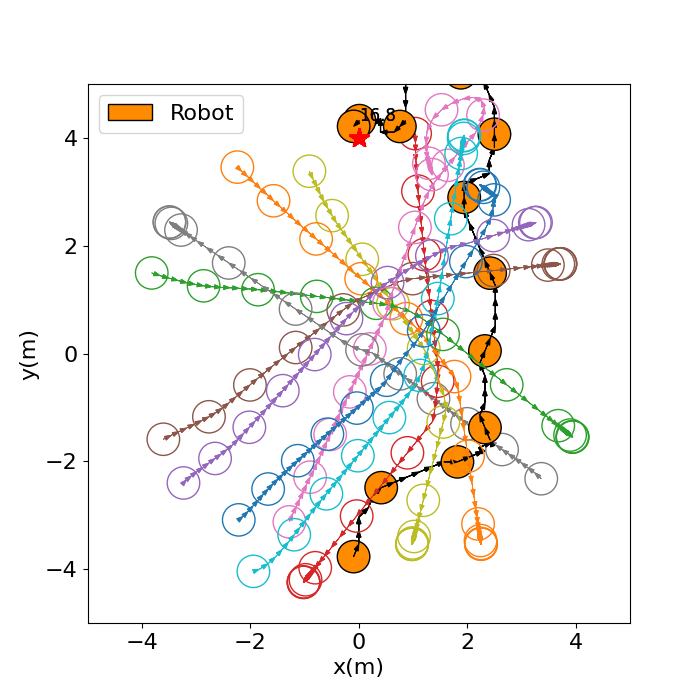}
\end{minipage}%
}
\subfigure[Ours]{
\begin{minipage}[b]{0.31\linewidth}
\centering
\includegraphics[width=\linewidth]{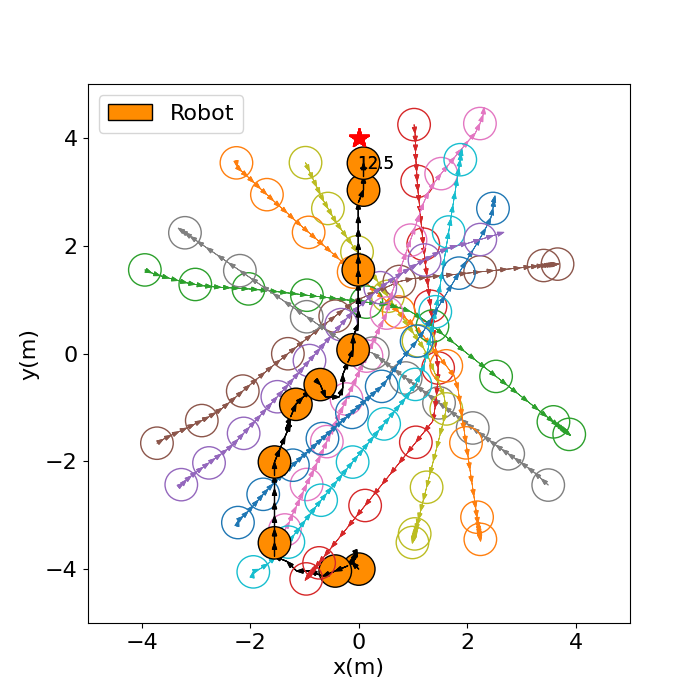}
\end{minipage}%
}%
\vspace{-3mm}

\subfigure[SARL]
{
\begin{minipage}[b]{0.31\linewidth}
\centering
\includegraphics[width=\linewidth]{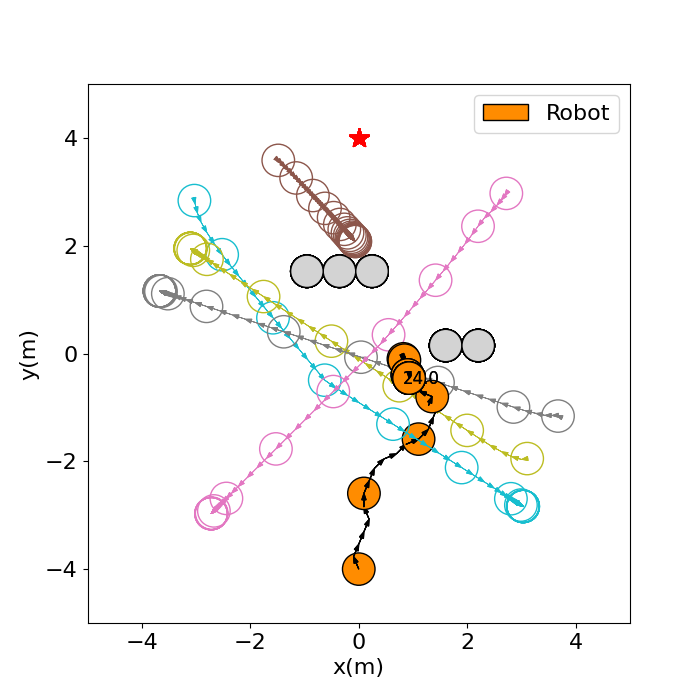}
\end{minipage}%
}
\subfigure[RGL]{
\begin{minipage}[b]{0.31\linewidth}
\centering
\includegraphics[width=\linewidth]{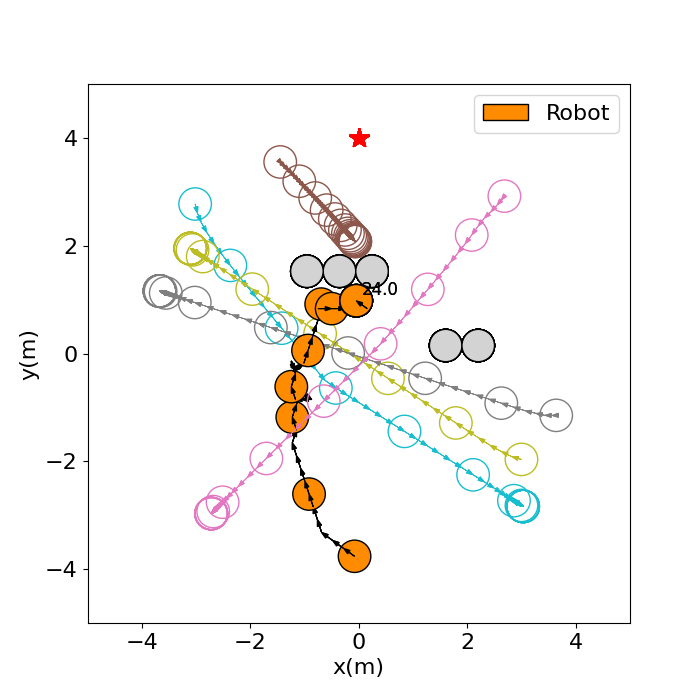}
\end{minipage}%
}
\subfigure[Ours]{
\begin{minipage}[b]{0.31\linewidth}
\centering
\includegraphics[width=\linewidth]{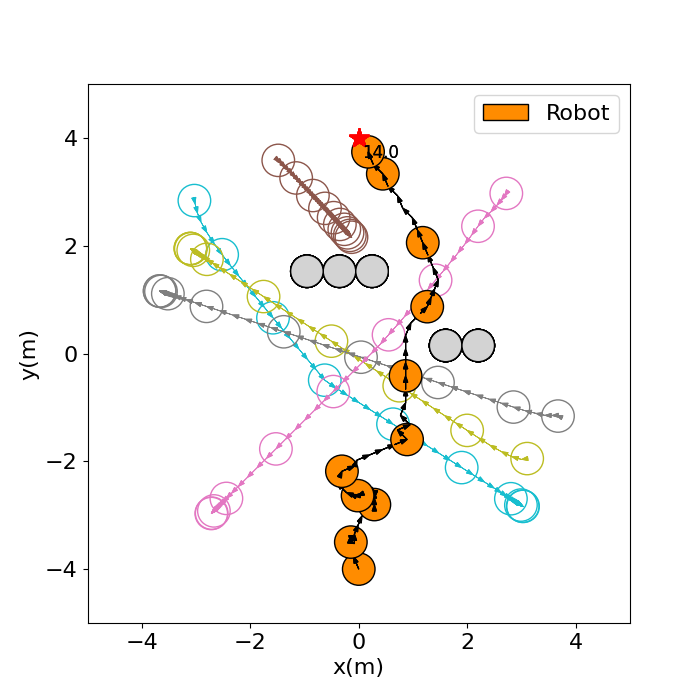}
\end{minipage}%
}%
\captionsetup{font={small}}
\caption{Trajectory comparisons under simple and complex scenarios. }
\label{fig:qualitative}
\end{figure}

\subsection{Ablation Study}

To assess the individual contributions of the different components in our model, we conduct an ablation study where we independently train and test the spatial and temporal branches with the same conditions as before. 
Fig.~\ref{fig:ablation} shows that the spatial graph and ASTG learn to navigate a safer side away, while the spatial graph has a longer navigation time. 
Furthermore, the temporal graph and ASTG are good at reasoning about the dynamics of the crowd, while ASTG demonstrates better social norm compliance.
These findings suggest that incorporating the spatial and temporal graph can significantly improve the performance and better adapt to dense scenarios with dynamic varying numbers of humans.

\subsection{Qualitative Evaluation}

To qualitatively evaluate the effectiveness of our method, we show in Fig.\ref{fig:qualitative} the robot's trajectories for different methods under simple and complex scenarios.
In the simple scenario with 10 dynamic humans, Fig.~\ref{fig:qualitative}(top), SARL collides with humans, while SG-DQN detours largely to achieve the goal.
In contrast, our method slightly detours at first and then directly reaches the goal, resulting in a shorter path.
When the scenarios are extended from simple to complex, Fig.\ref{fig:qualitative}(bottom) shows SARL and RGL have timeouts in the static group scenario.
As shown in Fig.\ref{fig:qualitative}(f), our method outperforms these two methods by slowing down at first and then going through the crowd. 
Spatial-temporal graph reasoning helps to understand relationships between agents over time and space and use that to make more informed decisions.
Our method thus is able to better adapt to dense and complex situations in uncertain environments.


\section{CONCLUSIONS}
\label{sec:con}
In this work, we proposed a novel attention-based spatial-temporal graph learning model for crowd robot navigation.
By formulating spatial and temporal graphs with a GAT, we implicitly compute both direct and indirect spatial interactions and temporal crowd features so that our model can reason for decision-making under changing scenarios.
Integrating a RNN, our model incorporates past trajectories into the temporal graph to also reason about the future intentions of each agent.
In the experiments, we showed our model outperforms baseline methods and can handle both uncertain simple and complex environments.
Based on the promising results, we intend in future work to further analyse and extend the GAT-module, for example using multiple graph layers, which can help to extract deeper information from the crowd.










{\small
\bibliographystyle{IEEEtran}
\bibliography{ref}
}

\end{document}